# A Hybrid Anomaly Detection Method for Electric Vehicle Identification from Smart Meter Recordings

**Ali Babalhavaeji[1], Hui Song[1], Mohammadreza Radmanesh[1], Mahdi Jalili[1]**

[1]RMIT University, Melbourne, Australia
Corresponding Author: Ali Babalhavaeji

**Abstract**--In recent years, electric vehicles (EVs) have attracted much attention due to their potential to decarbonize the transport system. EV charging imposes challenges on power grids, such as load imbalance and voltage instability. Therefore, electricity distributors need to identify which users have an EV. Smart meter recordings can be effectively used for this purpose. The major issue associated with such data is the imbalance problem, where the majority of the data belongs to users without EVs. The existing methods for EV identification from smart meter recordings have framed it as a classification task. In this paper, we frame the problem as an unsupervised problem and propose a hybrid anomaly detection method that combines the power of prediction-based and reconstruction-based anomaly detection methods. The proposed model considers temporal and stochastic representation information of smart meter time series via prediction-based and reconstruction-based components, respectively. The performance of the proposed model is compared with state-of-the-art supervised and unsupervised methods. Its superiority is verified on real-world data.

## I. Introduction

Using fossil fuels for energy demand results in greenhouse gas emissions, which leads to various climate change challenges [1]. To deal with climate change issues, the International Energy Agency (IEA) has set goals to restrain global temperature increase to two degrees Celsius. To that end, global CO2 emissions should reach zero by 2050 [2]. A big part of CO2 emissions, almost one quarter, is associated with transport systems. Electrifying the transport sector via electric vehicles(EVs) is a potential solution to decarbonize it and reach the emission reduction goals [3]. As an eco-friendly transportation system, EVs have attracted much attention in many countries [4].

The increasing uptake of EVs leads to additional load on the power grid, which results in challenges such as load unbalance, load peak shifting, voltage instability, and a negative impact on the power system [5]. Also, the power demand in the grid will change following the charging time and location [6]. Considering the lower energy cost and convenience of charging, many users choose to charge their EVs at home, especially during off-peak hours [7]. This raises economic and technical uncertainties, which affect the decision-making process for distribution system operators [8]. Electricity distributors cannot receive the EV purchase information on time, identifying new private EVs from the electricity consumption profiles is an essential task for them to better plan the network.

The problem of identifying EVs from smart meter recordings has been recently addressed in the literature. This is often modelled as a classification problem (two classes: users with and without an EV), where supervised learning methods are used to train appropriate classifiers [7, 9, 10]. The major issue associated with the problem of identification between EV users and non-EV users is the imbalance issue [11]. This problem is similar to a binary classification problem, where one class, the majority one, has many more samples in the sample space than the other one, the minority class [12]. This is the case in our problem, where the minority is EV users, and the majority are non-EV users. In a recent work by Hui et al [7], the problem is framed as a supervised learning problem. As the name suggests, these methods focus on preprocessing the training data instead of the model development process. They mostly take a resampling approach to deal with the imbalance issue. After resampling, the new training data is fed to classifiers like Support Vector Machine (SVM) [13], Random Forest (RF) [14], or more current and complex deep learning-based classifiers like Gated Recurrent Unit (GRU) [15]. The resampling approaches can be categorized into two major categories: oversampling and undersampling. The former refers to creating more instances from the minority class, and the latter refers to removing instances from the majority class [16]. Both these methods preprocess the data before sending it to the classifier. [7] takes an undersampling approach and proposes an ensemble classifier for EV identification. Andres et al [9] take a supervised approach and develop an RF. Volker et al [10] also take a supervised approach to solve the problem and develop a Convolutional Neural Network (CNN) and Long-Short Term Memory (LSTM).

Anomaly detection is another important problem that deals with imbalance issues. It is the case where the minority deviates significantly from the distribution of the majority [17]. Most anomaly detection methods take an unsupervised approach, which means they do not need to train the model with the minority data. They also don't need to resample the training data [18]. There are two major categories of anomaly detection methods: prediction-based and reconstruction-based. The former relies on temporal information of the data, and the latter relies on hidden structures of the data [19]. We have recently proposed using anomaly detection methods to identify energy users with EVs [20]. In this paper, we further extend our previous work and propose a hybrid anomaly detection approach to address the EV identification problem. We first frame our problem as an unsupervised anomaly detection problem, where EVs are considered anomalies. Then, we propose a hybrid anomaly detection method, which combines the power of a prediction-based anomaly detection method and reconstruction-based anomaly detection methods, to detect EV. The prediction-based method captures temporal information, and the reconstruction-based method captures stochasticity and representation information of the data. The contributions of our work are as follows:

- We frame the problem of EV identification from smart meter recordings as an unsupervised anomaly detection problem.
- We propose a hybrid anomaly detection method and apply it to the EV identification problem. It includes a prediction-based component and a reconstruction-based component.
- We explore the differences between the properties of data associated with EV users and non-EV users, which serves as a motivation to take an anomaly detection approach.
- We provide a comparison in terms of all important metrics and justify the reason behind falsely detected EVs and non-EVs.

The rest of this paper is organized as follows. In Section 2, we introduce the problem definition and the method. Section 3 includes data description and data analysis, results, and discussion. The results and discussion include the performance, ablation study, and discussion of why the proposed method outperforms other baselines. This paper is concluded in Section 4.

## II. Methodology

### A. Problem Definition

Our problem is to find energy users with an EV by analyzing electricity consumption time series. The EV charging load is not separated from the other appliances, and there is a single reading at each timestep, which is positive if the household imports electricity from the grid and negative if the household exports to the grid (some households have local generation from their rooftop solar). EV users are in the minority in Australia; therefore, anomaly detection methods can be used to detect them from those without EVs. We formulate our problem as an anomaly detection problem, where non-EV and EV users are considered as the normal and abnormal parts of the data, respectively. Let's denote the input with $\chi_t = (x_1, \cdots, x_t)$ and the output with $\varphi \in \{0,1\}$, where 0 indicates that the user does not have an EV.

For a sample to be an anomaly, it should have two main characteristics: i) it should be rare, and ii) deviate from normal behaviour. Based on the data analysis we conducted, we conclude that EV users' pattern of consumption has similar characteristics to anomalies, compared to non-EV users. Our problem becomes learning $\varphi: R^W \rightarrow \{1,0\}$ where 1 represents EV, 0 represents non-EV, and W is the length of the time window of the electricity consumption data. Based on the definition of the anomaly detection problem, we express the identification of EVs and non-EVs as follows:

$$\varphi(x) \rightarrow \begin{cases} 1 & if \varphi(x) > \tau \\ 0 & o.w \end{cases} \quad (1)$$

where $x \in \chi_t \subseteq R^W$ and $\tau$ is a threshold that determines the boundary between non-EV and EV users.

### B. Proposed Method

We propose a hybrid method based on prediction-based and reconstruction-based anomaly detection approaches. Time series data is a complex type of data, which has a stochastic nature as well. To consider all aspects of this type of data, we propose a hybrid method to consider temporal and representational information, while taking stochasticity into account as well.

Fig 1 shows the framework of the proposed model, called DHA (Deep Hybrid Anomaly detection). The first component preprocesses the time series of non-EV users to make time windows. Then, the windows are fed to the prediction-based and reconstruction-based components to extract temporal and stochastic representation features for each window. The prediction-based component uses Convolutional Neural Networks (CNN) and Gated Recurrent Unit (GRU) to extract temporal information. The reconstruction-based component uses CNN and variational inference to extract representation and stochasticity. The DHA keeps training on non-EV data until it reaches convergence.

After the training stage, we have a well-trained model on non-EV users, via which we produce a threshold $\tau$. The threshold serves as a boundary between EV users and non-EV users. We will delve more into the details of each stage in the next sections.

### C. *Training Phase*

#### 1) PREDICTION-BASED COMPONENT

This component uses two types of neural networks, CNN and GRU. The success of CNN in image processing is a good motivation to employ its capabilities for feature extraction in time series [21]. CNN usually includes three types of layers, which are convolutional, pooled, and fully connected layers. The convolutional layer aims to map the input features into a feature space with lower dimensions, called a feature map, using a sliding filter. It uses convolutional kernels (also known as filters). In this work, the input of the convolutional layer is a time window of the original features. The pooling layer creates a new representation using statistical operations on every element of the feature map. The statistical operation applies to the element and its neighbourhood. The model then passes the output of the pooling layer to the GRU component. The operation in the convolutional layer $l$ is formulated as follows:

$$y_l^j = \sum_{i\in K_j} w_l^{i,j} \otimes t_{l-1}^i + b_l^j \qquad (2)$$

$$t_l^j = Relu(y_l^j) \qquad (3)$$

where $t_l^j$ is $j$th feature map of the layer $l$, $K_j$ is the set of filters, $y_l^j$ is the output of the convolution operation, $w_l^{i,j}$ is the convolution kernel $i$ and $b_l^j$ is the bias. Each kernel(filter) creates a feature map. Then, a rectified linear activation function is used. In the pooling layer, we use the average operation [21].

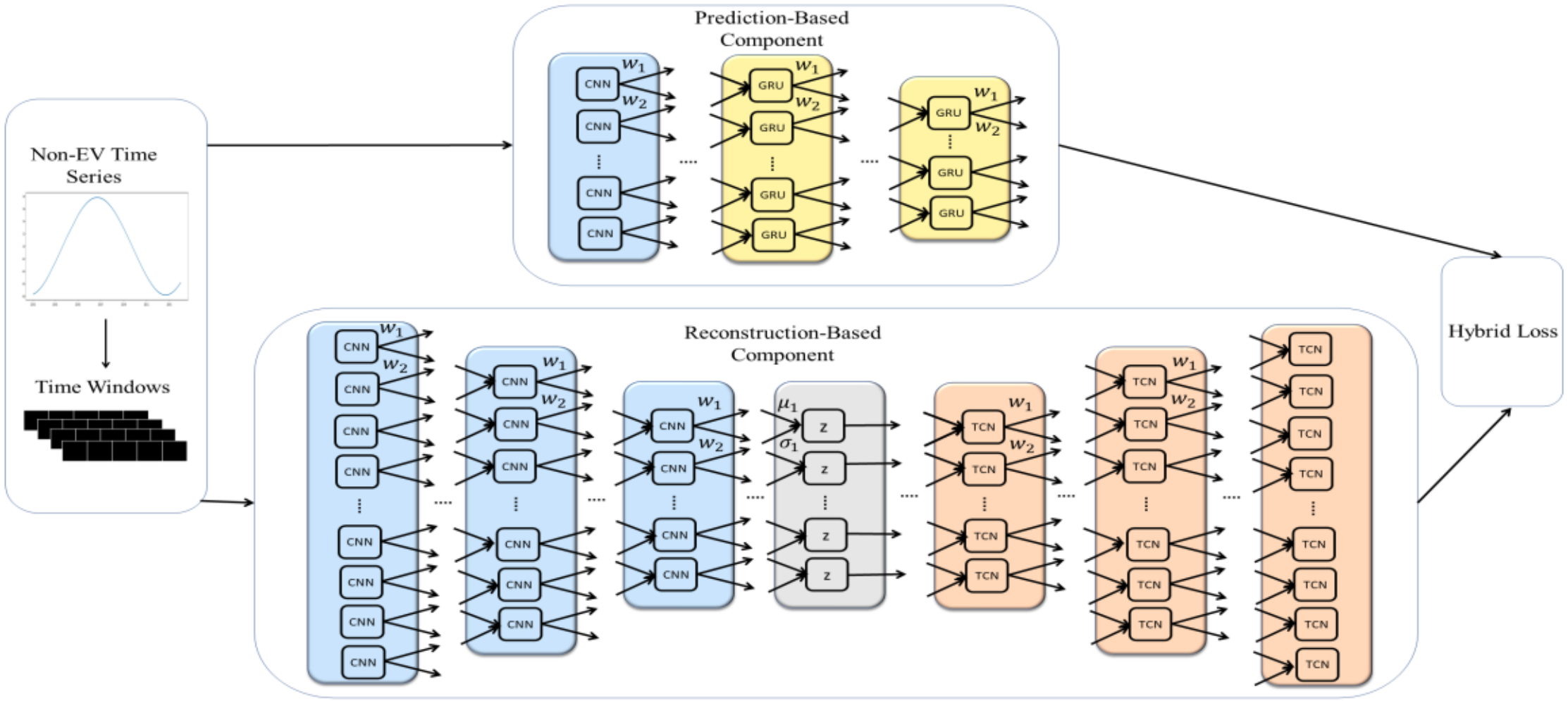

Fig 1: Framework of the DHA

After the feature engineering process in the CNN component, the prediction-based model uses GRU for extracting temporal information from the extracted feature maps. GRU and Long-Short Term Memory (LSTM) are the most popular variants of RNN models due to their capability in addressing gradient vanishing/exploding issues, which is a serious issue in vanilla RNNs. We base our development on GRU since it has fewer gates than LSTM. This leads to less complexity, which means the model will be less prone to overfitting. The GRU cells include two kinds of gates, the reset gate $r$ that determines whether the previous hidden state should be dropped, and the update gate $z$ that determines whether the output of the current state should be updated via $\tilde{h}$. We express the calculations as follows:

$$z_t = \sigma(w^z \cdot [h_{t-1}, x_t] + b^z) \qquad (4)$$

$$r_t = \sigma(w^r \cdot [h_{t-1}, x_t] + b^r) \qquad (5)$$

$$\tilde{h}_t = \tanh\left(w_{\tilde{h}_t} \cdot (h_{t-1} \odot r_t) + u_{\tilde{h}_t} x_t + b_g\right) \qquad (6)$$

$$h_t = z_t \odot h_{t-1} + (1 - z_t) \odot \tilde{h}_t \qquad (7)$$

where $z, r, \tilde{h}$ and $h$ are outputs of the update gate, reset gate, activation function, and output gate, respectively. $\sigma$ is the sigmoid activation function and $tanh$ is the hyperbolic tangent activation function. Finally, we minimize the mean squared loss function between the prediction and the real value as follows:

$$loss_{pre} = \|\chi_{t+1} - \hat{\chi}_{t+1}\|_2^2 \quad (8)$$

2) RECONSTRUCTION-BASED COMPONENT

In this component, we develop an autoencoder to capture the representational information of the time series. Our main goal is to learn a robust latent representation of non-EV users while considering the stochastic nature of time series data. The vanilla auto-encoders transform the input into a lower-dimension representation (a latent vector) and then reconstruct the input from this latent representation. An issue with the vanilla auto-encoders is that they might end up learning the identity function of the input. The other issue is that they encode each instance of the data independently of other instances, even if they belong to the same distribution. To learn a robust latent representation, we use convolutional networks as the encoder and convolutional transpose networks as the decoder. This makes the model robust to irregularities in the training data. To address the issues in vanilla auto-encoders, we use variational inference to force the latent representation to get close to a normal distribution as much as possible and to capture the stochasticity of the time series.

The variational process of DHA transforms the input $x$, via several 1d convolutional layers, into a latent stochastic variable $z$ with reduced dimensions. Then, it reconstructs $x$ from $z$, via 1d convolutional transpose layers. We assume that the input data are generated by a stochastic process. We express the marginal likelihood as the following integral:

$$p_\theta(x) = \int p_\theta(z) p_\theta(x|z)\, dz \quad (9)$$

where $p_\theta(z)$ is a prior distribution that we put on the latent variable, and $\theta$ is the parameter set. The computation of $p_\theta(x|z)$ leads to the computation of an intractable integral for $p_\theta(z|x)$ since the search space of $z$ is continuous and large. Therefore, we need to use an approximation. The function that approximates the integral is denoted by $q_\phi(z|x) \approx p_\theta(z|x)$, where $\phi$ is the parameter set. To make the approximation as close as possible to $p_\theta(z|x)$ , the negative of the Kullback-Leibler distance, $D_{KL}\left(q_\phi(z|x) \parallel p_\theta(z|x)\right)$ is included in the loss function. Overall, the encoding/inference part of the network, $E_\phi$, is responsible for computing the approximation posterior distribution of $q_\phi(z|x)$, and the decoder/generative part, $D_\theta$, is responsible for computing the conditional likelihood of $p_\theta(x|z)$. The network jointly optimizes the generative and inference parameters, $\theta$ and $\phi$. For the generative network, we use the reconstruction error in the loss function. The loss function for the joint optimization is expressed as follows:

$$loss_{rec} = E_{q_\phi(z|x)}(\log p_\theta(x|z)) + D_{KL}\left(q_\phi(z|x) \parallel p_\theta(z|x)\right) \quad (10)$$

For the first term, we need to sample $z$ first. This makes $z$ a random variable. Consequently, the sampling process makes the differentiation impossible. Without differentiation, the model cannot perform gradient descent to optimize the weights. To circumvent this issue, we use the reparameterization trick [22]. By doing so, we use the following transformation of the random variable of $z$:

$$z \sim N(\mu, \sigma) \rightarrow z = \mu + \varepsilon * \sigma \ ; \ \varepsilon \sim N(0,1) \quad (11)$$

3) THRESHOLD COMPONENT

This component uses the prediction and reconstruction losses over the training data to create the threshold $\tau$. $\langle X_{tr}, Y_{tr} \rangle$ is the training data where $X_{tr} = \{X_1, \dots, X_N\}$ and $Y_{tr} = \{Y_1, \dots, Y_N\}$. The DHA method uses a hybrid loss, which takes both temporal and stochastic representational information into account. We denote the threshold as follows:

$$\tau = \lambda loss_{rec}\left(X_{tr}, \hat{X}_{tr}\right) + (1-\lambda) loss_{pre}\left(Y_{tr}, \hat{Y}_{tr}\right) \quad (12)$$

where $\lambda$ is a trade-off hyperparameter between $loss_{rec}$ and $loss_{pre}$.

4) DETECTION PHASE

In this phase, DHA uses $\tau$ to do the detection. We have test data $D_{ts} = \langle X_{ts}, Y_{ts} \rangle$ where the data is preprocessed to make windows. DHA model derives the hybrid loss for each test data instance as follows:

$$E_{ts} = \left\{e_i = \lambda loss_{rec}\left(X_{ts}, \hat{X}_{ts}\right) + (1-\lambda) loss_{pre}\left(Y_{ts}, \hat{Y}_{ts}\right) \middle| \langle X_i^{ts}, Y_i^{ts} \rangle \in D_{ts}\right\} \quad (13)$$

Then, the detection function, denoted by $\varphi(X_i^{ts})$ is carried on as follows:

$$\varphi(X_i^{ts}) = \begin{cases} 0 & if\ e_i \le \tau \\ 1 & e.t \end{cases} \quad (14)$$

The DHA procedure is shown in Algorithm 1.

## III. Experiments

### D. Data Description

The dataset consists of electricity consumption time series for many customers with and without EVs in Victoria, Australia. The data is collected from March 2021 to May 2021 with a resolution of 30 minutes. It includes consumption time series of 1248 users (139 EV users and 1109 non-EV users). The EV brands consist of BMW, Tesla, Volvo, etc. Some of the users have wall chargers. Those who do not have any chargers can still charge their EV at home using standard wall brackets at a lower charging rate.

### E. Experimental Setup

We use Python 3.7.6 in Spyder 4.0.1 platform. We use Tensorflow 1.14.0 and the Keras library to develop the neural network. The configuration of baseline methods is explained in
Table 2.

**Algorithm1** Training and Detection Procedure of DHA

**Training Process**

**Input** Non-EV time series data $X$, batch size $M$, window size $K$
**Output** Threshold $\tau$, model parameters $W_{pre}$ and $W_{rec}$

```
Transform non-EV time series into windows X_tr = {X_1, ..., X_N}, and prediction targets Y_tr = {Y_1, ..., Y_N}
While not converged do:
Calculate feature maps for X_i through a 1d- convolutional network;// Error! Reference source not found., (3)
Calculate Y_i through GRU network with 64 and 32 neuron layers;// (4) and (7)
Calculate loss_pre, ∂loss_pre/∂w and update W_pre;// (8)
Calculate feature maps through 4 convolutional networks with 32, 64, 64, and 64 neurons;
Calculate parameters of the latent distribution, q_φ(z|x);
Sample z via reparameterization trick;// (11)
Reconstruct X_i via 4 layers of convolutional transpose layers;
Calculate the loss_rec, ∂loss_rec/∂w and update W_rec;// (10)
End while
Calculate τ;// (12)
Return τ, W_pre and W_rec
```

**Detection Process**

**Input:** test data, Threshold $\tau$
**Output:** Labels $\varphi(X_i)$ for all $X_i \in \{X_1, \dots, X_N\}$

```
Transform non-EV time series into windows X_ts = {X_1, ..., X_N}, and prediction targets Y_tr = {Y_1, ..., Y_N}
For X_i ∈ X_ts; i = 1, ..., N do:
Calculate Ŷ_i via the prediction-based network;
Calculate X̂_i via the reconstruction-based network;
Calculate hybrid loss ℒ_Xi = λ loss_rec(X_i, X̂_i) + (1 − λ) loss_pre(Y_i, Ŷ_i);
 IF ℒ_Xi > τ:
        φ(X_i) = "EV"
Else:
        φ(X_i) = "nonEV"
End for
Return Labels φ(X_i) for all X_i ∈ {X_1, ..., X_N}
```

TABLE 1: CONFIGURATION OF BASELINES

| | Baseline Methods for Comparison |
|---|---|
| GRU | First hidden layer: [5, 50], second hidden layer: [5, 30] |
| SVC | C: [0.1, 10000], $\gamma$: [0.0001, 1] |
| GB | Number of estimators: [5, 60], learning rate: [0.01: 1], the maximal number of features: [0.1, 1] |
| RF | Number of estimators: [5, 60], the maximal number of features: [0.1, 1] |
| DT | Maximal number of features: [0.1, 1], criterion: {gini; entropy}, minimal number of samples on each leaf: [2, 8] |
| LSTM | First hidden layer: [64], second hidden layer: [32] |
| CNNGRU | Number of Filters:[16], First Hidden Layer:[64], Second Hidden Layer:[32] |
| CNNLSTM | Number of Filters:[16], First Hidden Layer:[64], Second Hidden Layer:[1] |
| LOF | Number of neighbors:10 |
| IF | Number of estimators:100, contamination:0.01 |
| OCSVM | Kernel:"rbf", gamma:0.6 |
| LSTM-ED | Number of LTSM layer: 2, hidden size: 5, solver=Adam, learning rate (lr):1e-3 |
| DaGMM | Number of GMM= 2, MLP layer size={168,60,30,10,1}, solver=Adam, learning rate (lr):1e-3 |
| TCN | Loss={MSE}, Number of TCN blocks:1, kernel size:20, dilatation list:{1,2,4,8,16} |

We compare our method against baselines, including ECM [7], SVC [7], GRU [7], RF[9], LSTM [10], LSTM-ED [23], Deep auto-encoder with Gaussian Mixture Model (DaGMM) [24] TCN [25] (Table 1). Our baseline methods include both supervised and unsupervised approaches. The supervised methods use undersampling to provide EV labels for the training stage. ECM, SVC, GRU, RF, Decision Tree (DT) [7], LSTM, CNNGRU, and CNNLSTM are the supervised methods that take an undersampling approach. Local Order Factor (LOF), Isolated Forest (IF), OCSVM, LSTM encoder (LSTM-ED), and Temporal CNN (TCN) are unsupervised anomaly detection methods.

### *F. Evaluation Criteria*

To evaluate the performance of methods, we use four widely used metrics. The metrics are defined as follows:

$$Precision = \frac{TP}{TP + FP} \quad (15)$$

$$Recall = \frac{TP}{TP + FN} \quad (16)$$

$$F1 = 2 \times \frac{Precision \times Recall}{Precision + Recall} \quad (17)$$

$$Accuracy = \frac{TP + TN}{N} \quad (18)$$

where TP is the true positive, FP is the false positive, and FN is the false negative. In our case, true positive means EVs that are correctly detected as EVs, false positive means nonEVs that are falsely detected as EVs, and false negative means EVs that are falsely detected as nonEVs. $N$ is the size of the test data. In datasets with imbalance issues, F1 score is the most important metric.

### *G. Data Analysis*

An anomaly requires two main characteristics to be held for a part of a given dataset. First, it should be a rare part of the data. Second, it should deviate from the generation process that made up the majority of the data. We analyze the data that we collected to justify why anomaly detection methods should work for the detection of customers with an EV. The first condition holds since customers with EVs form a small part of the customer pool. For the second condition, we analyze the statistics of the data as follows. First, we compare the mean and standard deviation over all non-EV users and EV users. In Fig 2 We compare the mean active consumption between users with EVs and those without EVs over a window of 800 timestamps. It is observed that active consumption differs to a great extent. EV users sometimes reach more than 0.8 kWh of power consumption, while non-EV users reach a maximum of 0.5 kWh. Fig 3 shows a significant difference in standard deviation between EV and non-EV users. It shows that non-EV users have more similar consumption patterns than EV users. We have different categories of EV users according to their brands. Even one car brand might have several models. There is a high variance between EV users with different car brands. We observe big spikes for cars like 2x Tesla, Hyundai Kona, Jaguar, and Hyundai brands. Note that there is more than one model fo… we observe sig… ocorrelation. Fo… mp of a time se… ) and 80. All the… ong with the sc… n-EV part.

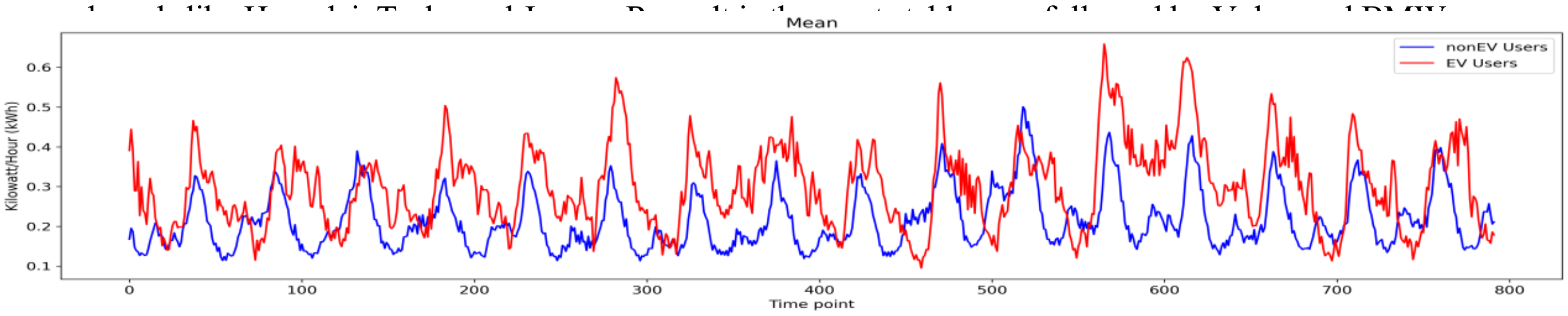

Fig 2: Comparison of Mean Consumption over non-EV and EV users (x-axis is a time window of 800 timestamps averaged over the whole dataset)

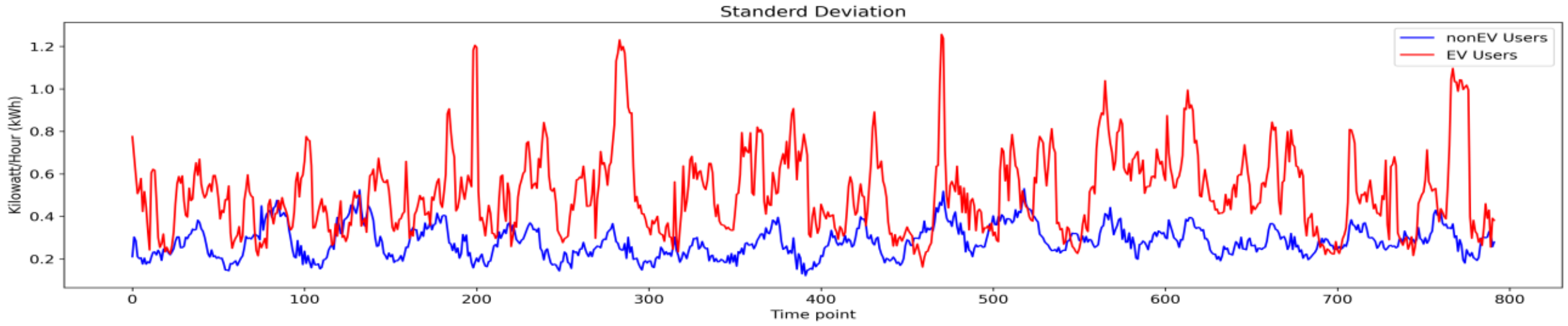

Fig 3: Comparison of Standard Deviation between non-EV and EV users

Table 2: Configuration of neural network models used in DHA.

| | Layers | Neurons | Activation Function |
|---|---|---|---|
| Prediction-Based Component | Zero Padding | | Linear |
| | 1D-Convolution | 32 | |
| | 1D-Pooling | | |
| | GRU 1 | 64 | Relu |
| | GRU 2 | 32 | Sigmoid |
| Reconstruction-Based Component | 4 layers of Zero Padding 1D-Convolution 1D-Pooling | 32, 64, 64, 64 | LeakyRelu |
| | 4 Layes of 1D-ConvolutionTranspose | 64, 64, 64, 32 | |

### H. Results & Discussion

We apply a grid search over the hyperparameters, including the number of neurons $\in \{16, 32, 64\}$, number of layers $\in \{1, 2, 3, 4\}$, the optimizer $\in \{Adam, SGD, RMSProp, Adagrad\}$, and the activation function $\in \{Linear, Relu, Sigmoid, Tanh\}$ for training the model.

Table 2 shows the final configuration of the neural network used for the training stage model.

Table 3 demonstrates that for almost all metrics, the proposed method (DHA) outperforms other baselines. As mentioned above, in cases where the imbalance issue holds, F1 score is the most important metric. DHA outperforms all baselines in terms of F2 score. DHA takes advantage of the availability of the normal population, which is the users without an EV in our case. This part of the data makes our model capable to learn the non-EV users' behaviour and deriving a good threshold to distinguish between EV and non-EV users. We use most of the non-EV part for the training module of DHA, and then we evaluate a mix of EV and non-EV data. We compare the performance of the proposed model with both supervised and unsupervised models. The supervised learning models use undersampling to balance the number of users in each class. For this undersampling, we use the method proposed in [7].

Regarding the comparison against supervised methods, DHA outperforms them by a great margin in terms of accuracy, precision, and F1 score. The accuracy of DHA is 0.8435. The second-best accuracy among the supervised methods is ECM method with an accuracy of 0.82. The best recall among supervised baselines is for RF with 0.8538, which suffers from a low precision of 0.3065, leading to a low F1 score of 0.4509.

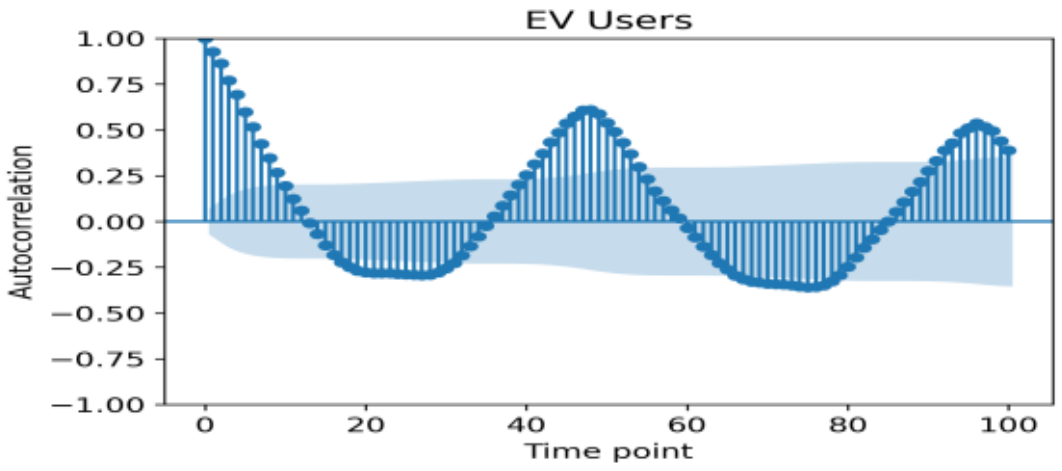

Fig 4: Autocorrelation analysis of EV users

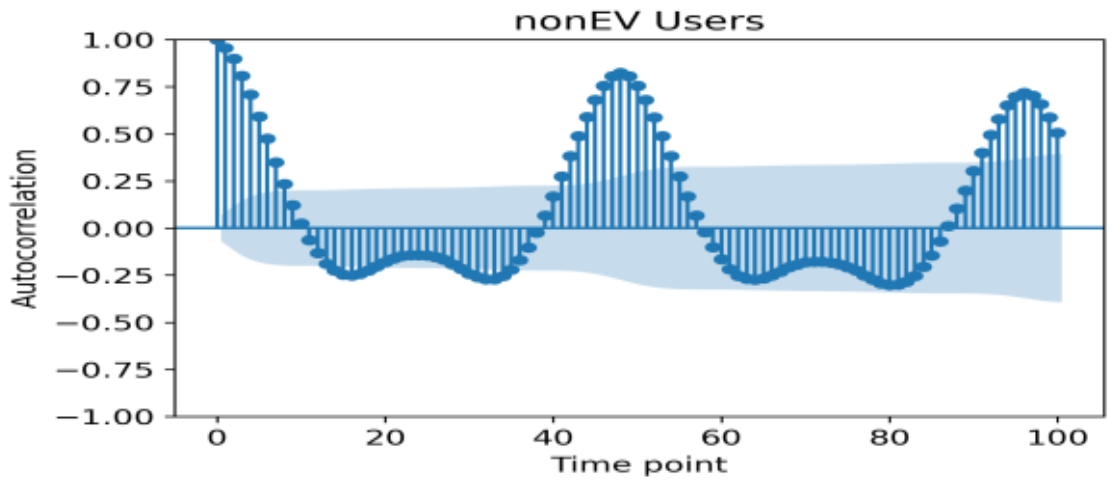

Fig 5: Autocorrelation analysis of nonEV users

The main purpose of the recall is to reward a method when it detects the target class correctly, i.e. true positive, and penalize it when it detects the target class falsely, i.e. false negative [26]. In our case, true positive means EVs that are detected truly as EVs, and false negative means EVs that are falsely detected as non-EVs. None of the supervised methods consider temporal information or stochasticity of the time series data, which results in a low F1 score of 0.4608 in the best case. In terms of precision, which penalized false positives, DHA reached 0.8688, whereas the best precision result among supervised methods reached a precision of 0.3374.

Other baselines are unsupervised anomaly detection methods, including LOF, IF, OCSVM, LSTM-ED, DaGMM,1D-CNN-AE, and TCN. LOF has the worst recall, 0.144, which results in the worst F1 as well. IF has the best recall, 0.9928, but it suffers from a high number of false positives, which manifest themselves in low precision. Therefore, IF has a low F1 of 0.2920. OCSVM suffers from a high number of false positives, which results in a low precision of 0.3333. Most of the baseline anomaly detection methods don't consider the temporal information of EV and non-EV consumption data, which leads to weaker performance than DHA. LSTM-ED considers the temporal information.

. Therefore, it better models the behavior of training data. LSTM-ED shows the second-best performance in terms of the F1 measure, i.e., 0.6667, followed by TCN, which also considers the temporal information and shows the F1 measure of 0.5949. 1D-CNN-AE is good at deriving the complex representation of data, but it does not consider temporal information. Therefore, it falls behind DHA, LSTM-ED, and TCN. DaGMM suffers from the same issue, although it is good at fitting Gaussian distribution to data. DHA has the best performance in detecting EVs and non-EVs. It is followed by LSTM-ED in this regard.

In terms of precision, DHA outperforms all baseline methods. The main goal of the precision measure is to reward correctly detected target classes and penalize falsely detected target classes. It means that, in contrast to recall, precision penalized false positives. DHA performs well in detecting EVs and non-EVs correctly. Good performance regarding precision often leads to good performance in terms of F1 measure as well. Therefore, the main advantage of DHA is fewer false positives compared to baselines, which is mainly due to the consideration of both temporal and stochastic representation information, and also eliminating the need for resampling. The CNN that we use in the training phase, in combination with the GRU component, makes the model capable to reduce noise, learn better features, considering the temporal information, and better learn the non-EV pattern in general. The use of a dropout layer combined with other regulations like the l1-norm prevents the model from overfitting. Variational inference also helps to consider the stochasticity of the time series.

Table 3: Performance in terms of Accuracy, Precision, Recall, and F1

| Supervised Baselines | Accuracy | Precision | Recall | F1 Score |
|---|---|---|---|---|
| RF | 0.7836 | 0.3065 | <u>0.8538</u> | 0.4509 |
| SVC | 0.812 | 0.3273 | 0.6923 | 0.4444 |
| GB | 0.7972 | 0.2905 | 0.6538 | 0.402 |
| DT | 0.6936 | 0.1966 | 0.6231 | 0.2981 |
| GRU | 0.6828 | 0.2445 | 0.7808 | 0.365 |
| LSTM | 0.7855 | 0.2694 | 0.6208 | 0.3757 |
| CNNGRU | 0.7406 | 0.2508 | 0.7524 | 0.3762 |
| CNNLSTM | 0.7063 | 0.2287 | 0.7690 | 0.3525 |
| ECM | 0.8224 | 0.3374 | 0.7308 | 0.4608 |
| Unsupervised Baselines | | | | |
| LOF | 0.2811 | 0.50 | 0.144 | 0.280 |
| IF | 0.2920 | 0.2942 | **0.9928** | 0.2920 |
| OCSVM | 0.3275 | 0.3333 | 0.719 | 0.1183 |
| LSTM-ED | <u>0.8247</u> | <u>0.6266</u> | 0.7122 | <u>0.6667</u> |
| DaGMM | 0.7061 | 0.4650 | 0.6691 | 0.5487 |
| 1D-CNN-AE | 0.6943 | 0.4873 | 0.6906 | 0.5714 |
| TCN | 0.7604 | 0.5311 | 0.6763 | 0.5949 |
| The proposed approach | | | | |
| DHA | **0.8425** | **0.8688** | 0.6347 | **0.7335** |

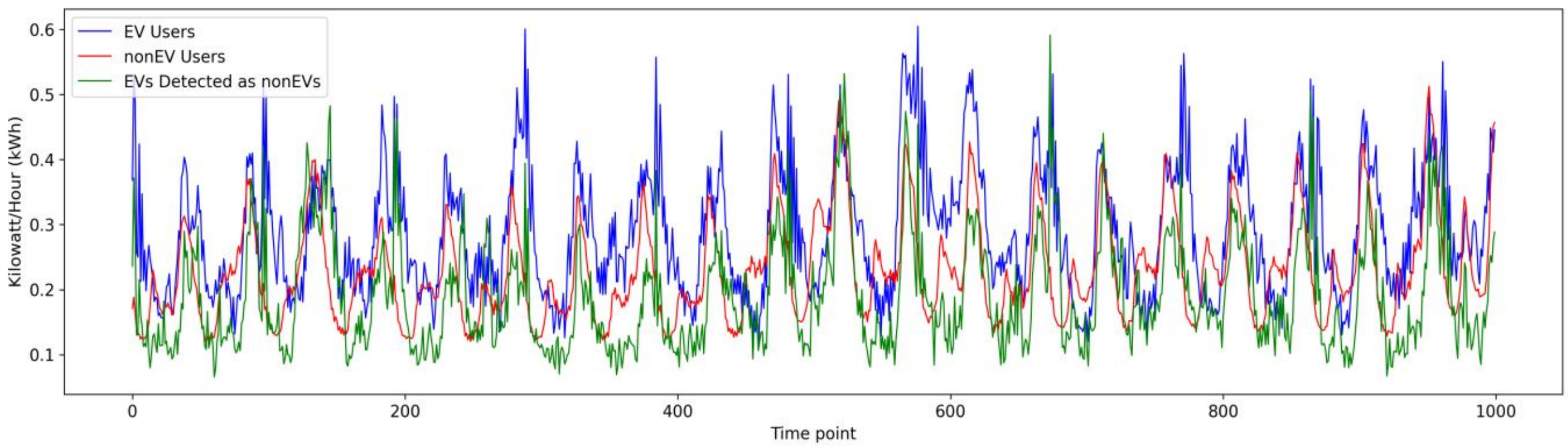

Fig 6: Comparison between Consumption pattern of EVs detected as non-EVs, with the pattern of average consumption associated with EV and non-EV Users (x-axis represents a window of 1000 timestamps averaged over the whole dataset)

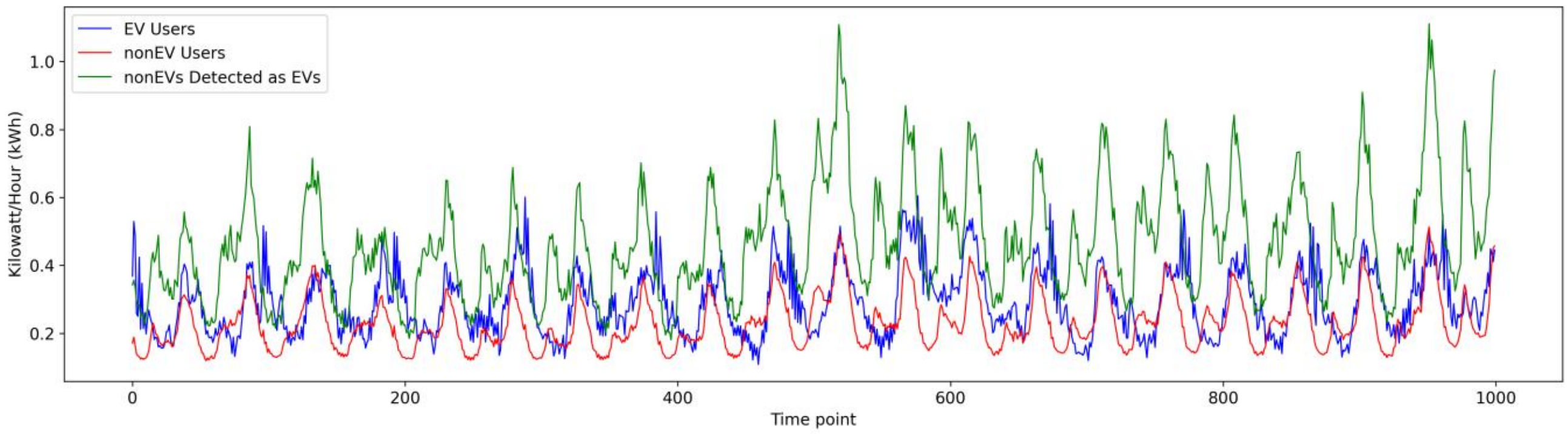

Fig 7: Comparison between the consumption pattern of non-EVs detected as EVs, with the pattern of average consumption associated with EV and non-EV users

There are still some false positives and false negatives detected by DHA. We delve more into the details of these instances. We compare the average consumption data of these users with the average pattern of the ground-truth EV and non-EV users. The results are depicted in Fig 6 and Fig 7. We observe in Fig 6 that the EV users, which are detected as non-EV users, have a pattern that is more like non-EV users. This must be the reason that our model detects them as non-EV users. In Fig 7, We see that non-EV users, which are detected as EV users, resemble EV user patterns more than non-EV users. In future work, we aim to work on this issue to boost the performance of DHA.

We also show the effectiveness of each component of DHA through the Ablation Study, which is the study of the contribution of each component in the final hybrid model. To do so, we create different versions of DHA, particularly pDHA, and rDHA. The former uses only the prediction-based component and drop reconstruction-based component, and the latter uses the reconstruction-based component only. We want to show the advantage of the combination of two components over the single components. Table 4 shows the comparison results. We observe that pDHA falls behind DHA in all metrics. It only considers the temporal information of the data and ignores representational information and the stochastic nature of the data. rDHA outperforms in terms of precision with 0.9921, which shows the capability of the reconstruction component and the stochastic model that we develop in capturing the patterns of non-EV data. It shows that rDHA has almost no false positives, and thus it detects almost all nonEVs correctly. But it falls behind in terms of the recall with 0.5526, indicating that rDHA suffers from many false negatives. It means that rDHA detects most EVs and non-EVs. On the other hand, DHA takes advantage of both temporal and representational information and captures the stochasticity of the data. That is why it outperforms in terms of the F1 measure. Therefore, we observe the advantage of the combination of two components over using them separately. This study shows that the combination of two prediction and reconstruction components helps to take advantage of both while compensating for each component's weaknesses.

Table 4: The ablation study shows the effectiveness of each component.

| Variants | Accuracy | Precision | Recall | F1 Score |
|---|---|---|---|---|
| pDHA | 0.7926 | 0.7801 | 0.6199 | 0.6846 |
| rDHA | 0.8360 | **0.9921** | 0.5226 | 0.6818 |
| DHA | **0.8425** | 0.8688 | **0.6347** | **0.7335** |

## IV. Conclusion

In this paper, we proposed a hybrid anomaly detection algorithm that combines the power of prediction-based anomaly detection approaches with reconstruction-based ones, for the EV identification problem from smart meter data. By doing so, we developed a method that captures temporal information, as well as representational and stochastic information of the time series associated with the consumption data. We first train the model on the consumption data of users without an EV and use a hybrid loss function. Then, we derived a threshold for non-EV consumption patterns using the hybrid loss function. Having the threshold and the trained model, we conducted the identification task on a mix of data from EVs and non-EV users. We proved empirically that our method outperforms both supervised and unsupervised methods. We also ran an ablation study to show the effectiveness of all components of our model. We observed the advantage of the combination of two components over using the components separately.